%% file: main.tex
\documentclass[sigconf]{acmart}
\usepackage{adjustbox}
\usepackage{booktabs}
\usepackage{graphicx}
\usepackage{subcaption}
\usepackage{multirow}
\usepackage{array}
\usepackage{amsmath}
\usepackage{enumitem}
\usepackage{algorithm}
\usepackage{algpseudocode}

\usepackage{tabularx, booktabs}
\usepackage{pgfplots}
\usepackage{tikz}
\usepackage{xcolor}
\usepackage{graphicx}
\usepackage{multirow}
\usepackage{pifont}
\usepackage{algpseudocode}
\usepackage{verbatim}

\pgfplotsset{compat=1.11,
    /pgfplots/ybar legend/.style={
    /pgfplots/legend image code/.code={%
       \draw[##1,/tikz/.cd,yshift=-0.25em]
        (0cm,0cm) rectangle (3pt,0.8em);},
   },
}
\definecolor{applegreen}{rgb}{0.55,0.71,0.0}

\AtBeginDocument{%
  \providecommand\BibTeX{{%
    \normalfont B\kern-0.5em{\scshape i\kern-0.25em b}\kern-0.8em\TeX}}}

\setcopyright{acmlicensed}
\copyrightyear{2018}
\acmYear{2018}
\acmDOI{XXXXXXX.XXXXXXX}

\acmConference[Conference acronym 'XX]{Make sure to enter the correct
  conference title from your rights confirmation emai}{June 03--05,
  2018}{Woodstock, NY}
\acmISBN{978-1-4503-XXXX-X/18/06}

\copyrightyear{2024}
\acmYear{2024}
\setcopyright{acmlicensed}\acmConference[KDD '24]{Proceedings of the 30th ACM SIGKDD Conference on Knowledge Discovery and Data Mining}{August 25--29, 2024}{Barcelona, Spain}
\acmBooktitle{Proceedings of the 30th ACM SIGKDD Conference on Knowledge Discovery and Data Mining (KDD '24), August 25--29, 2024, Barcelona, Spain}
\acmDOI{10.1145/3637528.3672035}
\acmISBN{979-8-4007-0490-1/24/08}
\begin{document}

\title{GAugLLM: Improving Graph Contrastive Learning for Text-Attributed Graphs with Large Language Models}

\author{Yi Fang}
\orcid{}
\affiliation{%
\department{SFSC of AI and DL}
\institution{New York University(Shanghai)}
\city{Shanghai}
\country{China}
}
\email{yf2722@nyu.edu}

\author{Dongzhe Fan}
\orcid{}
\affiliation{%
\department{SFSC of AI and DL}
\institution{New York University(Shanghai)}
\city{Shanghai}
\country{China}}
\email{df2362@nyu.edu}

\author{Daochen Zha}
\orcid{}
\affiliation{
  \department{Department of Computer Science}
  \institution{Rice University}
  \city{Huston}
  \country{USA}}
\email{daochen.zha@rice.edu}

\author{Qiaoyu Tan}
\orcid{}
\affiliation{%
\department{SFSC of AI and DL}
\institution{New York University(Shanghai)}
\city{Shanghai}
\country{China}}
\email{qiaoyu.tan@nyu.edu}

\renewcommand{\shortauthors}{Trovato and Tobin, et al.}

\begin{abstract}
This work studies self-supervised graph learning for text-attributed graphs (TAGs) where nodes are represented by textual attributes. Unlike traditional graph contrastive methods that perturb the numerical feature space and alter the graph's topological structure, we aim to improve view generation through \textbf{language supervision}. This is driven by the prevalence of textual attributes in real applications, which complement graph structures with rich semantic information. However, this presents challenges because of two major reasons. First, text attributes often vary in length and quality, making it difficulty to perturb raw text descriptions without altering their original semantic meanings. Second, although text attributes complement graph structures, they are not inherently well-aligned. To bridge the gap, we introduce GAugLLM, a novel framework for augmenting TAGs. It leverages advanced large language models like Mistral to enhance self-supervised graph learning. Specifically, we introduce a mixture-of-prompt-expert technique to generate augmented node features. This approach adaptively maps multiple prompt experts, each of which modifies raw text attributes using prompt engineering, into numerical feature space. Additionally, we devise a collaborative edge modifier to leverage structural and textual commonalities, enhancing edge augmentation by examining or building connections between nodes. Empirical results across five benchmark datasets spanning various domains underscore our framework's ability to enhance the performance of leading contrastive methods (e.g., BGRL, GraphCL, and GBT) as a plug-in tool. Notably, we observe that the augmented features and graph structure can also enhance the performance of standard generative methods (e.g., GraphMAE and S2GAE), as well as popular graph neural networks (e.g., GCN and GAT). The open-sourced implementation of our GAugLLM is available at \href{https://github.com/NYUSHCS/GAugLLM}{https://github.com/NYUSHCS/GAugLLM}.

\end{abstract}



\keywords{Graph contrastive learning, LLM for graph augmentation, Text-attributed graphs, Graph neural networks}



\maketitle

\section{Introduction}
\input{section/introduction}

\begin{figure*}[ht]
\centering
\includegraphics[width=15cm,height=6cm]{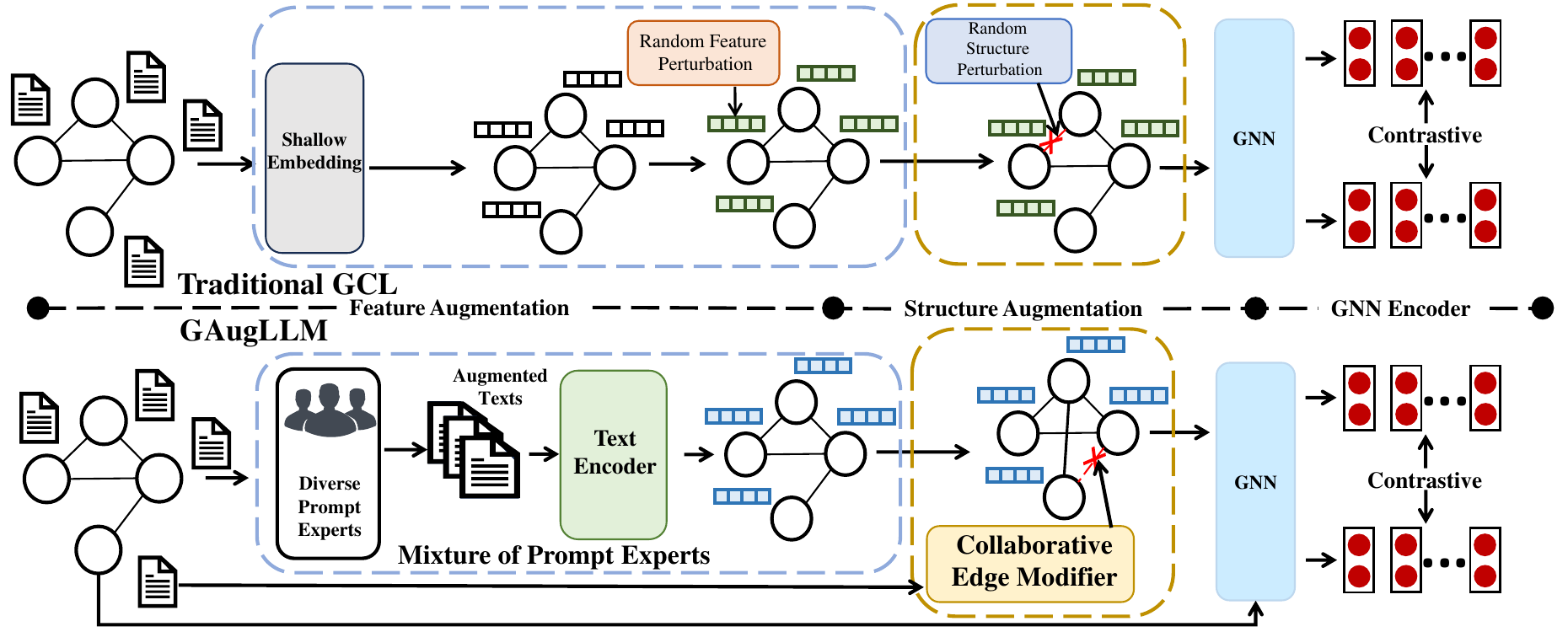}
\caption{The learning paradigm of GAugLLM vs. traditional GCL methods on TAGs. While standard GCL methodologies rely on text attributes primarily to generate numerical node features via shallow embedding models, such as word2vec, our GAugLLM endeavors to advance contrastive learning on graphs through advanced LLMs. This includes the direct perturbation of raw text attributes for feature augmentation, facilitated by a novel mixture-of-prompt-experts technique. Additionally, GAugLLM harnesses both structural and textual commonalities to effectively perturb edges deemed most spurious or likely to be connected, thereby enhancing structure augmentation.}
\label{fig:Figure-pipeline}
\vspace{-5pt}
\end{figure*}

\section{Related Work}
\input{section/relatedwork}

\section{Preliminary}
\input{section/preliminary}

\section{Methodology}
\input{section/method}

\section{Experiments}
\input{section/experiments}

\section{Conclusion and Future Work}
In this work, we delve into graph contrastive learning for text-attributed graphs (TAGs). While extensive endeavors have been proposed recently aimed at enhancing contrastive learning on graphs, these approaches are limited in harnessing the rich text attributes. This is because they simply utilize a shallow embedding model, such as word2vec, to transform the text attributes into feature space during pre-processing. To address this shortfall, we present GAugLLM, a pioneering graph augmentation framework that harnesses advanced LLMs for feature-level and structure-level augmentations. GAugLLM comprises two pivotal modules: the mixture-of-prompt-expert and collaborative edge modifier. The former dynamically integrates multiple prompt experts, each perturbing raw text attributes via prompt engineering, into the feature space for effective augmentation. The latter focuses on modifying connections in the original graph, either by deletion or addition, leveraging both structural and textual commonalities. Building upon these novel techniques, GAugLLM directly enhances the performance of leading contrastive learning methods (e.g., BGRL, GraphCL, and GBT). Interestingly, empirical findings indicate that GAugLLM can be readily applied to other GNN learning scenarios, including generative pre-training and supervised training. We hope our GAugLLM and experimental findings can motivate and pave the path for future research in leveraging LLMs for text-attributed graphs. In the future, we plan to extend GAugLLM to other graph-related tasks, such as graph generation, graph structure leanrning~\cite{zhou2024opengsl} and their applications in other domains.

\begin{acks}
The work is, in part, supported by Shanghai Frontiers Science Center of Artificial Intelligence and Deep Learning and the Startup fund at NYU Shanghai. 
\end{acks}

\balance 
\bibliography{reference}
\bibliographystyle{ACM-Reference-Format}

\section{Appendix}
\input{section/appendix}

\end{document}

%% file: section/introduction.tex
Graph data is ubiquitous across various domains, including traffic, e-commerce, chemistry, and bioinformatics. Unlike grid-like data such as images and text, graphs are non-Euclidean structures that capture intricate relationships between nodes, featuring diverse connection patterns. To address the complexities of graph data, Graph Neural Networks (GNNs) have emerged as specialized tools for representation learning~\cite{wu2020comprehensive,zhou2020graph,tan2019deep}. GNNs possess the capability to iteratively update node representations by aggregating information from neighboring nodes and themselves. Traditionally, the majority of GNN research has concentrated on supervised learning scenarios, where an ample amount of labeled graph data is available. 
However, annotating graph data is a laborious and expensive task. 

Consequently, recent attention~\cite{wu2020comprehensive} has shifted towards self-supervised graph learning, where the goal is to pre-train GNNs by generating training signals from unlabeled data itself. Once pre-trained, these models can serve as strong initializations for downstream supervised tasks with limited labeled samples~\cite{hassani2020contrastive,zhu2020deep,qiu2020gcc,gong2022ma,xu2021infogcl,suresh2021adversarial,tan2023collaborative,xia2022simgrace}, such as semi-supervised or few-shot learning scenarios. Graph contrastive learning (GCL), a prominent area in self-supervised graph learning, has shown remarkable effectiveness in pre-training GNNs~\cite{wu2020comprehensive}.
Existing GCL research, exemplified by GraphCL~\cite{you2020graph} and BGRL~\cite{thakoor2021large}, operate by creating two augmented views of the input graph and subsequently training GNN encoder to produce similar representations for both views of the same node. Various GCL methods~\cite{liu2022graph,xie2022self} differ in their designs for feature- and structure-level augmentation~\cite{ding2022data,hassani2020contrastive} and employ different contrastive learning objectives, e.g., InfoNCE~\cite{oord2018representation} and Barlow Twins~\cite{zbontar2021barlow}. 

Despite the numerous GCL methods proposed in recent years~\cite{velivckovic2018deep,zhu2021graph,hu2020gpt,zhang2021canonical,bielak2022graph}, they exhibit limitations when applied to graphs enriched with textual descriptions, often referred to as text-attributed graphs (TAGs). A typical example of TAGs is citation networks, where each node represents a research paper and includes text attributes like titles and abstracts.
These text attributes offer valuable information for enhancing graph learning due to their expressiveness, capturing intricate semantic nuances. However, previous GCL efforts simply utilize textual attributes to derive numerical features using shallow embedding models such as Word2vec~\cite{mikolov2013efficient} or Bag-of-Words (BoW)~\cite{harris1954distributional}. Subsequently, they perform feature-level perturbation on this transformed feature space. While conceptually simple, this feature augmentation strategy is inherently suboptimal. It cannot fully capture the complexity of semantic features~\cite{he2023explanations,chen2023exploring}, and the quality of augmented features is constrained by the text transformation function used. Furthermore, these methods perform structure augmentation in an attribute-agnostic manner, relying solely on stochastic perturbation functions like edge masking. Nevertheless, as previously discussed in~\cite{lee2022augmentation,gong2023ma,xia2022simgrace}, randomly perturbing edges in the original graph can be risky. Therefore, text attributes represent a valuable resource to advance graph augmentations for effective contrastive learning. 

However, leveraging text attributes for effective graph augmentation presents several challenges. \textbf{Firstly, maintaining original semantic meanings while performing text augmentation is difficult,} as text attributes in real-world graphs often vary in length and quality (see Table~\ref{dataset_stat}). Traditional heuristic augmentation strategies, such as random word replacement, insertion and swap, may be sub-optimal in such cases. \textbf{Secondly, mapping augmented text attributes into numerical space poses another challenge.} Unlike traditional GCL methods that transform text data into feature vectors in the pre-processing step, directly perturbing input text attributes requires a principled text transformation function capable of capturing the disparity between augmented and original text attributes. Moreover, this transformation function should be personalized w.r.t. each node, as nodes in a graph often exhibit different characteristics. \textbf{Thirdly, augmenting topological structure solely based on text attributes is ineffective and inefficient,} due to the heterogeneity of text attributes and graph structure. While an intuitive solution is to estimate edge weights between nodes by calculating their similarity in the text space and generating an augmented graph by sampling over the edge space using estimated edge weights, this approach suffers from scalability issues. The complexity is quadratic to the graph size, which could be millions or even billions in practice. Moreover, it may lead to a sub-par augmented graph with connection patterns significantly different from the original graph topology since text attributes and graph structure are not well aligned in general. Hence, an effective structure augmentation strategy should jointly consider both text attributes and the original graph structure.

To fill this research gap, in this work, we present GAugLLM, a novel graph augmentation framework for self-supervised learning on graphs. The key idea is to utilize advanced large language models (LLMs), such as Mistral and LLaMa, to perturb and extract valuable information in the text space, enabling effective feature- and structure-level augmentation. Specifically, to address the first two challenges, we introduce a mixture-of-prompt-expert technique to perturb original text attributes based on diverse prompt experts, each representing a specific prompt template tailored to an LLM. Subsequently, a smaller LLM (e.g., BERT) is fine-tuned to dynamically integrate multiple augmented text attributes into the feature space. This transformation considers node statistics and adopts observed node connections as training supervision. To tackle the third challenge, we propose a collaborative edge modifier strategy. This approach reduces augmentation complexity by prioritizing the most spurious and likely connections between each node and others from a structural perspective. Then an LLM is adopted to identify the most promising connections in the context of text attributes. Overall, our main contributions are summarized below:
\begin{itemize}[leftmargin=*, topsep=0mm]
\item We introduce a novel graph augmentation approach, namely GAugLLM, designed for text-attributed graphs. Unlike standard GCL methods that solely transform text attributes into feature vectors and conduct feature- and edge-level perturbation independently, GAugLLM leverages rich text attributes with LLMs to jointly perform perturbation in both feature and edge levels. 
\item We propose a mixture-of-prompt-expert method to generate augmented features by directly perturbing on the input text attributes. Unlike heuristic-based random perturbation, we utilize powerful LLMs to disturb text attributes from diverse prompt aspects, which are then dynamically integrated into a unified feature space as augmented features. 
\item We devise a collaborative edge modifier scheme to leverage text attributes for structural perturbation. Unlike traditional edge perturbation functions, e.g., random masking, we offer a principled approach that adds and deletes node connections by jointly looking at the textual and structural spaces. 
\item We extensively experiment on various TAG benchmarks across different scales and domains to validate the effectiveness of GAugLLM. Our empirical results demonstrate that GAugLLM improves the performance of leading contrastive methods (e.g., BGRL, GraphCL, and GBT), with up to 12.3\% improvement. Additionally, we consistently observe gains by utilizing the augmented features and structures of our model on popular generative methods (e.g., GraphMAE and S2GAE) and graph neural networks (e.g., GCN and GAT).
\end{itemize}

%% file: section/relatedwork.tex
Our work is closely related to the following two directions. Readers, who are interested in GNNs and LLMs, please refer to~\cite{yang2023harnessing} and~\cite{min2023recent,zhao2023survey} for a comprehensive review. 

\noindent\textbf{Self-supervised learning on graphs.} Self-supervised learning has become a compelling paradigm for learning representations from graph-structured data without explicit annotations. The existing work can be mainly divided into two categories: contrastive learning methods and generative methods. Contrastive learning approaches learn graph representations by maximizing the similarity between positive pairs while minimizing the similarity between negative pairs. Previous research, such as GraphCL~\cite{you2020graph}, has further advanced contrastive learning methods by introducing various graph data augmentation techniques. These methods generally rely on effective strategies for positive and negative sample pairing and robust Graph Neural Network (GNN) architectures to extract graph features. More recently, GPA~\cite{zhang2024graph} provides personalized augmentation methods for for graphs. Generative methods focus on learning graph representations by predicting unseen parts of the graph. For instance, S2GAE~\cite{tan2023s2gae} masks edges in the graph and predicts missing links, while GraphMAE~\cite{hou2022graphmae} utilizes GNN models as the encoder and decoder to reconstruct masked node features. Recently, GiGaMAE~\cite{shi2023gigamae} learns more generalized and comprehensive knowledge by considering embeddings encompassing graph topology and attribute information as reconstruction targets. Generative methods encourage the model to capture the intrinsic structure and evolution patterns of graphs, leading to richer and more insightful graph representations.

\noindent\textbf{Representation learning on TAGs.} 
Text-attributed graphs have recently received significant attention in both academia and industry. Initially, representation learning on TAGs relied on shallow embedding methods. Although these approaches provided a foundation for representation learning on TAGs, they are limited by their inability to deeply integrate text and graph structure information. 
GIANT~\cite{chien2021node} represents a leap forward by more effectively integrating deep textual information with graph topology. By doing so, GIANT can capture complex dependencies and interactions between text and structure, significantly improving performance on downstream tasks. Recently, some studies have been focused on leveraging the sophisticated capabilities of LLMs to enhance the understanding and analysis of TAGs. TAPE~\cite{he2023explanations} leverages LLMs for generating explanations as features, which then serve as inputs for graph neural networks (GNNs), thereby enriching the representation of TAGs. GLEM~\cite{zhao2023learning} proposes a novel approach that combines GNNs and LMs within a variational Expectation-Maximization (EM) framework for node representation learning in TAGs. However, they mainly focus on supervised training.


%% file: section/preliminary.tex
In this section, we introduce notations, formalize the research problem of this work, and 
illustrate prospective opportunities for harnessing language models to enhance contrastive learning on TAGs.

\noindent\textbf{{Text-Attributed Graphs.}} We are given a TAG $\mathcal{G}=\{\mathcal{V}, \mathcal{S},\mathbf{A}\}$ with $N$ nodes, where $\mathcal{V}$ denotes the node set, and $\mathbf{A}\in\mathbb{R}^{N\times N}$ represents the adjacency matrix. For each node $v\in\mathcal{V}$ is associated with a textual attribute $S_v$, and $\mathcal{S}=\{S_v|v\in\mathcal{V}\}$ is the attribute set. 

In this work, we study self-supervised learning on TAGs. Specifically, the goal is to pre-train a mapping function $f_\theta: \mathcal{S}\times \mathbf{A}\rightarrow \mathbb{R}^d$, so that the semantic information in $\mathcal{S}$ and the topological structure in $\mathbf{A}$ could be effectively captured in the $d$-dimensional space in a self-supervised manner.  

\noindent\textbf{{Graph Neural Networks.}} For graph-structure data, graph neural networks (GNNs) are often applied to instantiate $f_\theta$. Specifically, the goal of GNNs is to update node representation by aggregating messages from its neighbors, expressed as: 
\begin{equation}
\mathbf{h}^{(k)}_v=\text{COM}(\mathbf{h}^{(k-1)}_v, \text{AGG}(\{\mathbf{h}_{u}^{(k-1)}: u\in\mathcal{N}_v\})),
\label{eq:mps}
\end{equation}
where $\mathbf{h}_v^{(k)}$ denotes the representation of node $v$ at the $k$-th layer and $\mathcal{N}_v=\{u|\mathbf{A}_{v,u}=1\}$ is a direct neighbor set of $v$. In particular, we have $\mathbf{h}_v^{(0)}=\mathbf{x}_v$, in which $\mathbf{x}_v=\text{Emb}(S_v)\in\mathbb{R}^F$ is a $F$-dimensional numerical vector extracted from $v$'s textual attribute $S_v$ and $\text{Emb}(\cdot)$ stands for embedding function. The function \text{AGG} is used to aggregate features from neighbors~\citep{kipf2016semi}, and function \text{COM} is used to combine the aggregated neighbor information and its own node embedding from the previous layer~\citep{vaswani2017attention}.

\noindent\textbf{{Graph Contrastive Learning on TAGs.}} 
Let $\tau_f: \mathbb{R}^F \xrightarrow{} \mathbb{R}^F$ and $\tau_s: \mathcal{V} \times \mathcal{V} \xrightarrow{} \mathcal{V} \times \mathcal{V}$ represent the feature-level and structure-level perturbation functions, respectively. An example of $\tau_f$ is feature masking~\citep{jin2020self}, while for $\tau_s$, edge masking~\citep{zhu2021empirical} serves as a typical illustration. Previous GCL endeavors~\citep{you2021graph,yin2022autogcl,zhang2022graph} typically start by employing a shallow embedding function $g: S\rightarrow \mathbb{R}^F$, such as Word2vec and BoW, to transform text attributes into numerical feature vectors, i.e., $\mathbf{x}_v=g(S_v)$ as a preprocessing step. Subsequently, they generate two augmented graphs, $\mathcal{G}_1 = ({\mathbf{A}_1, \mathbf{X}_1})$ and $\mathcal{G}_2 = ({\mathbf{A}_2, \mathbf{X}_2})$, by applying perturbation functions to the transformed feature space $\mathbf{X}$ and graph structure $\mathbf{A}$. Here, $\mathbf{X}_1 = \{\tau_f^1(\mathbf{x}_v)|v\in\mathcal{V}\}$, $\mathbf{A}_1 = \tau_s^1(\mathbf{A})$, $\mathbf{X}_2 = \{\tau_f^2(\mathbf{x}_v)|v\in\mathcal{V}\}$, and $\mathbf{A}_2 = \tau_s^2(\mathbf{A})$. Then, two sets of node representations are acquired for the two views using a shared GNN encoder, denoted as $\mathbf{H}_1$ and $\mathbf{H}_2$, respectively. Finally, the GNN encoder is trained to maximize the similarity between $\mathbf{H}_1$ and $\mathbf{H}_2$ on a node-wise basis. In this study, we mainly focus on three state-of-the-art methods, namely GraphCL~\citep{you2020graph}, BGRL~\citep{thakoor2021large}, and GBT~\citep{bielak2022graph}, for experimentation. 

\noindent\textbf{Opportunity.} Existing GNN studies have been restricted in their utilization of text attributes, which are both informative and valuable in TAGs~\cite{yan2023comprehensive}. First, the shallow embedding function $g$ is limited in its ability to comprehend the semantic information of text attributes, particularly when compared with LLMs like Mistral and LLaMa. Second, it is well understood that node attributes and graph structure are complementary to each other~\cite{liao2018attributed,huang2017label}. Therefore, merely perturbing the graph structure without considering their semantic similarity may result in a suboptimal augmented graph, whose semantic meaning diverges significantly from the original structure~\cite{lee2022augmentation}. Motivated by the above opportunities for improvement, in this work, we explore the following research question: \textit{Can we leverage text attributes to enhance the performance of graph contrastive learning from the perspective of graph augmentation?}  


%% file: section/method.tex
\begin{figure}[t]
\centering
\includegraphics[width=8.6cm]{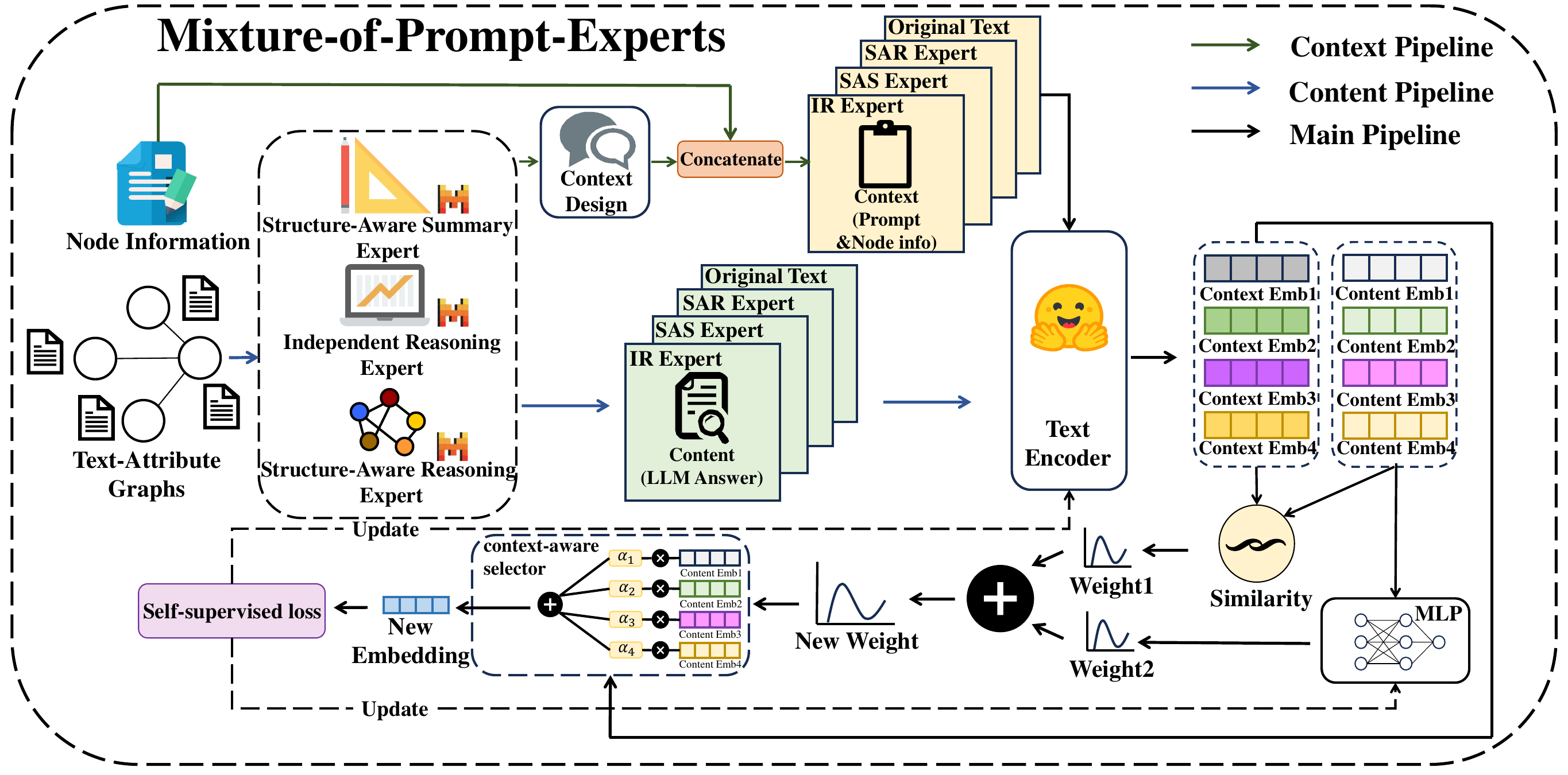}
\caption{The pipeline of the mixture-of-prompt-experts for feature augmentation. It takes a TAG as input and then utilizes multiple prompt experts to perturb the original text attributes, generating diverse augmented attributes. These augmented text attributes are then integrated into a unified augmentation feature by considering the graph statistics as attention context. }
\label{fig:Figure-mope}
\vspace{-0.5cm}
\end{figure}

In this section, we present the proposed GAugLLM shown in Figure~\ref{fig:Figure-pipeline}. We first discuss how to perturb raw text attributes for effective feature augmentation (in Section~\ref{sec-mope}). Then, we elaborate on a tailored collaborative edge modifier to effectively add or delete edges for structure augmentation (in Section~\ref{sec-cem}). Finally, we show how the proposed feature- and structure-level augmentation strategies can be extended to the standard GCL pipeline (in Section~\ref{sec-apply}). 

\subsection{Mixture-of-Prompt-Experts}
\label{sec-mope}
As discussed above, traditional GCL methods are limited in leveraging rich text attributes for feature augmentation, as they solely rely on a shallow embedding model to transform text attributes into the feature space during a pre-processing step. These transformed features are then fed into a perturbation function $\tau_s$ for feature perturbation. To make full use of text attributes for feature augmentation, we propose a novel framework called mixture-of-prompt-experts. 

Figure~\ref{fig:Figure-mope} depicts the overall architecture, which offers an elegant approach to directly perturb text attributes and map them into the feature space. Given a TAG $\mathcal{G}=(\mathcal{V},\mathcal{S},\mathbf{A})$ as input, our model initially perturbs the text attribute $S_v$ of node $v$ into diverse augmented texts ($\{\hat{S}_v^i\}_{i=1}^m$) using different prompt experts $\{f_{pe}^i\}_{i=1}^m$, where $m$ represents the number of total experts. Let $f_{\Theta_{\text{text}}}$ denote the text transformation function with parameters $\Theta_{\text{text}}$, and $\hat{\mathbf{x}}_{v}^i$ indicate the hidden embedding of the $i$-th augmented text produced by $f_{pe}^i$.

\subsubsection{Prompt experts.} Our mixture-of-prompt-experts approach begins by configuring a diverse set of prompt experts to perturb the raw text attribute $S_v$ while preserving its semantic meanings. Motivated by the remarkable success of LLMs (e.g., LLaMA and Mistral) in understanding and generating natural language, we initialize our prompt experts with LLM yet with different prompt designs. Specifically, we design three different prompt templates to perturb the raw text attributes from the structural and reasoning perspectives, as illustrated below.  

\begin{itemize}[leftmargin=*, topsep=0mm]
\item \textbf{Structure-Aware Summarization (SAS Expert).} Let $\mathcal{S}_v^N=\{S_u|v\in \mathcal{N}_v\}$ represent the textual attribute set of node $v$'s neighbors. The idea of SAS is to query the LLM to create a summary of the anchor node $v$ by comprehending the semantic information from both its neighbors and itself. 
The general prompt format is illustrated in Figure~\ref{graph-augmentation}. 

\item \textbf{Independent Reasoning (IDR Expert).} In contrast to SAS, which concentrates on text summarization, IDR adopts an ``open-ended'' approach when querying the LLM. This entails instructing the model to make predictions across potential categories and to provide explanations for its decisions. The underlying philosophy here is that such a reasoning task will prompt the LLM to comprehend the semantic significance of the input textual attribute at a higher level, with an emphasis on the most vital and relevant factors~\cite{he2023explanations}. 
The general prompt format is illustrated in Figure~\ref{graph-augmentation}. 

\item \textbf{Structure-Aware Reasoning (SAR Expert).} Taking a step beyond IDR, SAR integrates structural information into the reasoning process. The rationale for this lies in the notion that connected nodes can aid in deducing the topic of the anchor node. 
The general prompt format is given in Figure~\ref{graph-augmentation}. 

\end{itemize}

Based on the three prompt experts, we can map the text attribute $S_v$ of each node $v$ into three augmented texts $\{\hat{S}_v^{i}|i\in\{\text{SAS},\text{IDR},\text{SAR}\}\}$

\subsubsection{Text encoder.} After perturbing the raw text attributes, we need to train a text encoder mapping the augmented texts into hidden space. Instead of using shallow embedding algorithm, we aim to fine-tune a smaller LLM (e.g., BERT) to encode the domain-specific text data. In particular, given the augmented text set $\{\hat{S}_v^i|i\in\{\text{SAS},\text{IDR},\text{SAR}, \text{Raw}\}\}$ of node $v$, the text encoder works as follows:
\begin{equation}
    \hat{\mathbf{x}}_v^{i} = f_{\Theta_{\text{text}}}(\hat{S}_v^i),
\end{equation}
where $\mathbf{x}_v^i\in\mathbb{R}^D$ denotes the feature vector of the $i$-th prompt expert produced by the text encoder. Therefore, for each node $v$, we can generate four augmented feature vectors in total, each representing one prompt expert accordingly. Notably, we include the raw text attribute as the fourth prompt expert inspired by~\cite{zhu2021graph,you2021graph}. 

\subsubsection{Context-aware selector.} Given the $m$ initial augmented feature vectors $\{\hat{\mathbf{x}}_v^i\}_{i=1}^m$ of node $v$, the next question is how to select the most relevant one for each node. As discussed in study~\cite{you2021graph}, different graphs may benefit from different types of augmentation strategies. Similarly, each prompt expert can be seen as as specific perturbation strategy. Therefore, an intuitive solution is to employ an attention mechanism to dynamically integrate the most relevant expert by computing attention coefficients, formulated as: 
\begin{equation}
    \alpha_{v}^i=\frac{exp(\mathbf{W}_1\hat{\mathbf{x}}_v^i/\tau)}{\sum_{k=1}^{m}exp(\mathbf{W}_1\hat{\mathbf{x}}_v^k/\tau)},
    \label{alo-atten1}
\end{equation}
where $\mathbf{W}_1\in\mathbb{R}^{1\times D}$ denote the trainable attention weights, and $\alpha_v\in\mathbb{R}^{m}$ is the attention vector for node $v$. $\tau$ is the temperature parameter used to adjust the sharpness of the attention distribution.  

While effective, Eq.~\eqref{alo-atten1} neglects the node statistics when integrating various prompt experts. To address this, we introduce the notion of \textit{context prompt}, which describes the functionality of each prompt expert and the node statistics, such as degree information. We report the context prompt for different prompt experts in Appendix~\ref{sec-feat-augmentor}. Let $S_v^{(c,i)}$ denote the context prompt of node $v$ for the $i$-th prompt expert, we calculate the context-aware attention distribution of node $v$ as follows: 
\begin{equation}
    \alpha_{v}^{c,i}=\frac{exp(f_{\Theta_{\text{text}}}(S_v^{(c,i)})\mathbf{W}_2\hat{\mathbf{x}}_v^i/\tau)}{\sum_{k=1}^{m}exp(f_{\Theta_{\text{text}}}(S_v^{(c,k)})\mathbf{W}_2\hat{\mathbf{x}}_v^k/\tau)}.
    \label{alo-atten2}
\end{equation}
$\alpha_v^c\in\mathbb{R}^{m}$ is context-aware attention vector for node $v$, $\mathbf{W}_2\in\mathbb{R}^{D\times D}$ is the model weights. Eq.~\eqref{alo-atten2} offers the flexibility to incorporate both node-level and prompt expert-level prior knowledge into the attention process. Finally, we integrate the two attention mechanisms and rewrite Eq.~\eqref{alo-atten1} as:

\begin{equation}
    \alpha_{v}^i=\frac{exp((\mathbf{W}_1\hat{\mathbf{x}}_v^i+f_{\Theta_{\text{text}}}(S_v^{(c,i)})\mathbf{W}_2\hat{\mathbf{x}}_v^i/\tau))}{\sum_{k=1}^{m}exp((\mathbf{W}_1\hat{\mathbf{x}}_v^k+f_{\Theta_{\text{text}}}(S_v^{(c,i)})\mathbf{W}_2\hat{\mathbf{x}}_v^k/\tau))},
    \label{alo-atten3}
\end{equation}
Based on Eq.~\eqref{alo-atten3}, we obtain the final augmented feature vector $\hat{\mathbf{x}}_v$ of node $v$ as: $\hat{\mathbf{x}}_v=\sum_{i}\alpha_v^i\hat{\mathbf{x}}_v^i$. 

\noindent\textbf{Training objective.} To effectively fine-tune the pre-trained smaller LLM ($f_{\Theta_{\text{text}}}$) within our text attribute space, we train $f_{\Theta_{\text{text}}}$ to reconstruct the observed connections. Specifically, given node $v$ and its corresponding row in the adjacency matrix $\mathbf{A}_{v,:}$, we frame the fine-tuning task as a multi-label classification problem. However, directly fine-tuning $f_{\Theta_{\text{text}}}$ on a high-dimensional output space of size $|\mathcal{V}|$ is computationally infeasible. To address this challenge, we employ the extreme multi-label classification (XMC) technique used in GAINT~\cite{chien2021node} for efficient optimization. 

\subsection{Collaborative Edge Modifier}
\label{sec-cem}
Up to this point, we have discussed the process of obtaining augmented feature vectors $\{\hat{\mathbf{x}_v}\}$ using text attributes. Now, we will explore how text attributes can be utilized for effective structure perturbation. In essence, the aim of edge perturbation is to enhance the diversity between the original and augmented structures while maintaining their structural patterns. In our context, edge perturbation faces two major hurdles: 1) the quadratic growth of the edge search space relative to the graph size, resulting in huge computational costs when querying LLM; 2) the semantic disparity between the text space and observed topological structure, making it suboptimal to rely solely on one of them for edge perturbation. 

To tackle this challenge, we propose a text-aware edge perturbation framework, called collaborative edge modifier. As outlined in Algorithm~\ref{algo-cem} of Appendix~\ref{appen-edge}, it leverages the commonalities between both data modalities for edge perturbation. The first stage involves structure-aware top candidate generation. Specifically, we adopt a standard network embedding algorithm (e.g., DeepWalk) to map nodes into a hidden space using only structure data. Subsequently, we assess the similarity between any two nodes based on their network embeddings. For each node $v$, we then create two disjoint edge sets $\mathcal{E}_v^{\text{spu}}$ and $\mathcal{E}_v^{\text{mis}}$. The former contains the top $K$ least similar edges among the observed links, representing the most spurious connections. The latter comprises top $K$ most similar edges among the disconnected links in the original graph, indicating likely/missing connections. 

After obtaining the two candidate sets $\mathcal{E}_v^{\text{spu}}$ and $\mathcal{E}_v^{\text{mis}}$ of node $v$, the second stage aims to modify the two sets using text attributes. In particular, we define a simple edge modifier prompt to query LLM determining whether two nodes should be connected by interpreting their semantic similarity. The detailed template for this prompt is reported in Section~\ref{appen-edge} of the Appendix. Let $S_{v,u}$ denote the query prompt for nodes $v$ and $u$, we define the addition and deletion operations below. 

\subsubsection{Edge deletion.} This operation is designed for the potential spurious set $\mathcal{E}_v^{\text{spu}}$. We ask the LLM to estimate the likelihood of each edge $e\in \mathcal{E}_v^{\text{spu}}$ using corresponding query prompt, resulting in an action sequence $a^{\text{del}}_v\in\mathbb{R}^{|\mathcal{E}_v^{\text{spu}}|}$. Here, $a^{\text{del}}v(i)=1$ if the LLM believes the two nodes should be disconnected and $a^{\text{del}}_v(i)=0$ otherwise. 

\subsubsection{Edge addition.} In addition to edge deletion, we also define the addition operation to add potential missing links in $\mathcal{E}_v^{\text{mis}}$. We query the LLM to assess the likelihood of each edge $e\in \mathcal{E}_v^{\text{mis}}$ using the corresponding query prompt, leading to an action sequence $a^{\text{add}}_v\in\mathbb{R}^{|\mathcal{E}_v^{\text{mis}}|}$. $a^{\text{add}}_v(i)=1$ if the LLM believes the two nodes should be connected; $a^{\text{add}}_v(i)=0$ otherwise.

\noindent\textbf{Remark.} The two stages offer a principled approach to determining the connections between two nodes based on structural and textual aspects, leveraging the commonalities of the two modalities. Furthermore, by focusing on the two action sets $\mathcal{E}_v^{\text{spu}}$ and $\mathcal{E}_v^{\text{mis}}$, the potential query space on the LLM is significantly reduced from the complexity of $O(|\mathcal{V}|^2)$ to $O(K)$. $K$ is a hyperparameter, such as 10 in practice. In summary, the output of the proposed collaborative edge modifier is a set of action sequences $\{a_v|v\in\mathcal{V}\}$, where $a_v=a_v^{\text{del}}||a_v^{\text{add}}$ and $||$ stands for concatenation operation. It is worth noting that this process is conducted ``off-the-fly''.  

\begin{table*}[t]
\centering
\caption{Dataset statistics of five text-attributed graphs (TAGs).}
\label{dataset_stat}
\begin{tabular}{c!{\vrule width \lightrulewidth}c|c|c|c!{\vrule width \lightrulewidth}c!{\vrule width \lightrulewidth}c!{\vrule width \lightrulewidth}c} 
\toprule
Data                                       & \# Nodes & \# Edges    & \# Features & \# Classes & \# Average Text & \# Longest Text & \# Shortest Text  \\ 
\hline
PubMed                                     & $19,717$ & $44,338$    & $500$       & $3$        & $1649.25$             & $5732$                 & $18$                     \\ 
\hline
Ogbn-Arxiv                                 & $169343$ & $1,166,243$ & $128$       & $40$       & $1177.993$            & $9712$                 & $136$                    \\ 
\hline
\multicolumn{1}{c|}{Books-History}         & $41,551$ & $400,125$   & $768$       & $12$       & $1427.397$            & $103130$               & $27$                     \\ 
\hline
\multicolumn{1}{c|}{Electronics-Computers} & $87,229$ & $808,310$   & $768$       & $10$       & $492.767$             & $2011$                 & $3$                      \\ 
\hline
\multicolumn{1}{c|}{Electronics-Photo}     & $48,362$ & $549,290$   & $768$       & $12$       & $797.822$             & $32855$                & $5$                      \\
\bottomrule
\end{tabular}
\end{table*}

\begin{table*}[t]
\centering
\caption{Semi-supervised accuracy results of state-of-the-art GCL methods advanced. "SE" denotes the feature matrix obtained by shallow embedding models. "GIANT" indicates that the text transformation is implemented by the method proposed in ~\cite{chien2021node}. }
\label{table-GAugLLM-results}
\begin{tabular}{l|l|l|l|l|l|l} 
\toprule
                           & Method  & BGRL                & GBT                 & GraphCL              & GCN                   & GAT                           \\ 
\midrule
\multirow{3}{*}{PubMed}    & SE      & 80.6±1.0(+3.60\%)   & 79.44±1.31(+5.34\%) & 79.8±0.5(+2.79\%)    & 77.8±2.9(+3.59\%)     & 78.7±2.3(+0.88\%)             \\
                           & GIANT   & 82.75±0.28(+0.91\%) & 81.13±0.82(+3.14\%) & 81.21±0.22(+1.01\%)~ & 79.32±0.45(+1.60\%)   & 78.80±0.52(+0.75\%)           \\
                           & GAugLLM & \textbf{83.50±0.84} & \textbf{83.68±1.90} & \textbf{82.03±1.74}  & \textbf{80.59±0.82}   & \textbf{79.39±1.13}           \\ 
\hline\hline
\multirow{3}{*}{Arxiv}     & SE      & 71.64±0.12(+2.89\%) & 70.12±0.18(+1.68\%) & 70.18±0.17(+1.23\%)  & 71.74 ± 0.29(+2.58\%) & 71.59±0.38(+2.18\%)           \\
                           & GIANT   & 73.14±0.14(+0.78\%) & 70.66±0.07(+0.91\%) & 70.94±0.06(+0.15\%)  & 73.29±0.10(+0.41\%)   & \textbf{74.15±0.05(-1.34\%)}  \\
                           & GAugLLM & \textbf{73.71±0.08} & \textbf{71.3±0.18}  & \textbf{71.05±0.14}  & \textbf{73.59±0.10}   & 73.15±0.05                    \\ 
\hline\hline
\multirow{3}{*}{Photo}     & SE      & 57.98±0.09(+31.8\%) & 68.56±0.95(+14.0\%) & 53.21±0.47(+36.3\%)  & 60.31±0.71(+26.7\%)   & 59.03±0.59(+28.6\%)           \\
                           & GIANT   & 71.65±0.61(+6.64\%) & 74.65±0.69(+4.72\%) & 71.40±0.62(+1.55\%)  & 71.83±0.38(+6.35\%)   & 71.44±0.49(+6.27\%)           \\
                           & GAugLLM & \textbf{76.41±0.64} & \textbf{78.17±0.54} & \textbf{72.51±0.78}  & \textbf{76.39±0.62}   & \textbf{75.92±0.42}           \\ 
\hline\hline
\multirow{3}{*}{Computers} & SE      & 69.53±0.26(+20.5\%) & 70.67±0.54(+14.6\%) & 53.51±0.27(+51.7\%)  & 59.43±0.90(+41.5\%)   & 58.17±0.67(+43.7\%)           \\
                           & GIANT   & 74.23±0.56(+12.3\%) & 76.87±0.36(+5.37\%) & 74.24±0.24(+8.88\%)  & 76.72±0.22(+9.61\%)   & 75.63±0.49(+10.5\%)           \\
                           & GAugLLM & \textbf{83.8±0.34}  & \textbf{82.74±0.45} & \textbf{80.83±0.36}  & \textbf{84.10±0.20}   & \textbf{83.60±0.18}           \\ 
\hline\hline
\multirow{3}{*}{History}   & SE      & 69.84±0.42(+9.29\%) & 71.62±0.38(+6.27\%) & 57.26±0.44(+32.2\%)  & 58.14±1.76(+33.1\%)   & 66.39±0.82(+17.65\%)          \\
                           & GIANT   & 74.16±0.83(+2.93\%) & 71.89±0.63(+5.90\%) & 71.14±0.38(+6.45\%)  & 75.99±0.10(+1.87\%)   & 74.67±0.39(+3.44\%)           \\
                           & GAugLLM & \textbf{76.33±0.88} & \textbf{76.11±0.4}  & \textbf{75.73±0.35}  & \textbf{77.41±0.32}   & \textbf{78.11±0.52}           \\
\bottomrule
\end{tabular}
\end{table*}

\subsection{Graph Contrastive Learning for TAGs}
\label{sec-apply}
Given the augmented feature matrix  $\hat{\mathbf{X}}$ and the set of edge perturbations $\{a_v|v\in\mathcal{V}\}$, we can enhance the performance of existing GCL methods by replacing their augmentation strategies with ours. Specifically, prior studies aim to maximize the mutual information between two augmented views, denoted by $(\mathbf{A}_1, \mathbf{X}_1)$ and $(\mathbf{A}_2,\mathbf{X}_2))$. Now we can pre-train a GNN encoder to maximize the mutual information between $(\mathbf{A},\mathbf{X})$ and $(\hat{\mathbf{X}},\hat{\mathbf{A}})$. Here, $\mathbf{X}$ is the feature matrix obtained based on raw text attributes, i.e., $\mathbf{X}_v=f_{\Theta_{\text{text}}}(S_v)$, and $\hat{\mathbf{A}}$ is constructed by random sampling (e.g., with uniform distribution) some actions from $\{a_v|v\in\mathcal{V}\}$ in a simple wise fashion per iteration. Notably, due to the randomness in edge action selection, the augmented views $(\hat{\mathbf{X}},\hat{\mathbf{A}})$ will vary across different iterations, albeit in a consistent manner thanks to the definition of these action sequences. Additionally, as the augmented feature matrix $\hat{\mathbf{X}}$ builds upon the original text attributes, it is generally more effective than $\mathbf{X}$ and can incentivize the GNN encoder to learn more valuable textual information.

In addition to GCL methods, we have observed that our model could also be extended to enhance the performance of other popular graph generative models (e.g., GraphMAE and S2GAE), as well as standard GNN methods such as GCN and GAT, simply by leveraging the augmented features and structures as input. We empirically analyze this applicability in Section~\ref{sec-overal}.

%% file: section/experiments.tex
Throughout the experiments, we aim to address the following research questions. \textbf{RQ1:} Can GAugLLM enhance the performance of standard graph contrastive learning methods? \textbf{RQ2:} How does GAugLLM perform when applied to other GNN learning scenarios, such as generative pre-training and supervised learning? \textbf{RQ3:} How does each component of GAugLLM, i.e., different prompt templates of mixture-of-prompt-experts, attention mechanism, and the collaborative edge modifier, contribute to the performance? \textbf{RQ4:} Is the proposed collaborative edge modifier sensitive to the random sampling process in each iteration?

\subsection{Experimental Setup} 
\textbf{Datasets.}\ \ We evaluate the proposed GAugLLM framework using five publicly available TAG datasets. These datasets encompass two citation networks, namely PubMed~\cite{sen2008collective} and Ogbn-Arxiv (Arxiv)~\cite{hu2020open}, and three E-commerce datasets extracted from Amazon~\cite{ni2019justifying}, including Electronics-Computers (Compt), Books-History (Hist), and Electronics-Photography (Photo). For all of these datasets, we adhere to the standard data splits used in prior research. In our experiments, we opt to utilize raw texts directly rather than processed text features so that the textual semantics are preserved. The statistical details of these datasets are outlined in Table~\ref{dataset_stat}.

\textbf{Baselines.}\ \ We compare GAugLLM with two textual feature extraction methods. \textbf{Shallow Embedding (SE)} is the standard way of generating textural features with shallow embedding models (i.e., Word2vec~\cite{mikolov2013efficient} or Bag-of-Words (BoW)~\cite{harris1954distributional}). SE serves as the baseline result of a GCL or GNN algorithm. \textbf{Graph Information Aided Node feature exTraction (GIANT)~\cite{chien2021node}} is a state-of-the-art graph-agnostic feature extraction algorithm tailored for raw texts in graphs. It fine-tunes a language model with self-supervised learning and then fuses the textual embedding with the graph structure information to make predictions.


\textbf{Experimental Details.}\ \ 
We conduct experiments upon three state-of-the-art GCL methods, namely GraphCL~\citep{you2020graph}, BGRL~\citep{thakoor2021large}, and GBT~\citep{bielak2022graph}, and two standard GNNs methods: GCN~\citep{kipf2016semi} and GAT~\citep{velivckovic2018graph}. For the reproducibility of our experiments, we employ GNN implementations from the PyG~\citep{fey2019fast} package. For the GraphCL, BGRL, and GBT methods,
we closely adhere to the procedures outlined in~\cite{zhu2021empirical}. For each experiment, we run 5 times and report the mean result and the standard deviation. By default, we use the open-sourced LLM model -- Mixtral 8*7b version. We provide detailed experimental configurations in Section~\ref{appen-config} of Appendix. 

\subsection{Overall Evaluation} 
\label{sec-overal}
To answer \textbf{RQ1}, We conduct extensive experiments on five benchmark TAG datasets in standard semi-supervised node classification tasks. Table~\ref{table-GAugLLM-results} presents the results for three popular GCL backbones and two standard GNN methods. From these results, we make the following observations.

\begin{table}[t]
\centering
\caption{Accuracy results of generative methods on TAGs.}
\label{table:generative-results}
\begin{tabular}{l|l|l|l} 
\toprule
                           & Method  & S2GAE                & GraphMAE             \\ 
\midrule
\multirow{3}{*}{pubmed}    & SE      & 81.66±1.32           & 81.1±0.4             \\
                           & GIANT   & 82.43±0.61           & 80.16±0.08           \\
                           & GAugLLM & \textbf{83.02±0.94 } & \textbf{82.98±0.77}  \\ 
\hline\hline
\multirow{3}{*}{arxiv}     & SE      & 68.38±0.13           & 71.75±0.11           \\
                           & GIANT   & \textbf{70.91±0.09}  & 72.58±0.15           \\
                           & GAugLLM & \textbf{71.23±0.08}  & \textbf{73.4±0.13}   \\ 
\hline\hline
\multirow{3}{*}{Photo}     & SE      & 76.12±0.75           & 67.49±0.59           \\
                           & GIANT   & \textbf{77.89±0.48}  & 71.66±0.48           \\
                           & GAugLLM & 76.77±0.22           & \textbf{74.11±0.37}  \\ 
\hline\hline
\multirow{3}{*}{Computers} & SE      & 82.70±0.27           & 70.90±0.38           \\
                           & GIANT   & \textbf{84.37±0.42}  & 73.91±0.17           \\
                           & GAugLLM & \textbf{84.32±0.36}  & \textbf{78.57±0.3}   \\ 
\hline\hline
\multirow{3}{*}{History}   & SE      & 71.80±0.82           & 71.77±0.24           \\
                           & GIANT   & 73.56±0.92           & 75.59±0.62           \\
                           & GAugLLM & \textbf{74.84±1.02}  & \textbf{76.84±0.33}  \\
\bottomrule
\end{tabular}
\end{table}

~\ding{172} \textbf{GAugLLM can significantly boost the performance of state-of-the-art GCL methods across all datasets.} In Table~\ref{table-GAugLLM-results}, GAugLLM consistently outperforms SE and GIANT across all 15 testing scenarios (i.e., columns of BGRL, GBT, and GraphCL). Specifically, while GAINT performs notably better than the SE method due to its utilization of a smaller LLM for transforming text attributes into the feature space, GAugLLM surpasses GAINT in all cases. This superiority can be attributed to the advantage of the proposed mixture-of-prompt-experts, which augments the raw text attributes from diverse aspects. Notably, GAugLLM achieves improvements of +20.5\% and +12.3\% over SE and GIANT, respectively, when training BGRL on the Computers dataset. Moreover,~\ding{173} \textbf{GCL methods generally outperform standard GNNs when using different textual feature extractors}. This is expected because GCL methods have the potential to learn superior representations and effectively utilize unlabeled data. Our GAugLLM further enhances the learned representations of GCL methods by more effectively encoding textual information into the model. These results demonstrate the effectiveness of GAugLLM in harnessing rich textual features. 


In addition to the contrastive learning scenario, we also test the applicability of the learned augmented features on other GNN learning settings, such as generative pre-training and supervised learning (\textbf{RQ2}). Table~\ref{table-GAugLLM-results} and Table~\ref{table:generative-results} summarize the results on supervised GNN methods and generative pre-training methods, respectively. We observed that~\ding{174} \textbf{GAugLLM is primarily designed for enhancing GCL, it also significantly improves the performance of standard GNN methods.} In the last two columns of Table~\ref{table-GAugLLM-results}, GAugLLM consistently outperforms SE in all testing cases and surpasses GIANT in 9 out of 10 testing scenarios. Particularly on the Computers dataset, GAugLLM outperforms the standard GAT and GAT+GIANT by +43.7\% and +10.5\%, respectively. This strong performance can be attributed to the utilization of a mixture of prompt experts, which enable the incorporation of informative textual semantics enhanced by advanced LLM into model training, thereby benefiting various GNN methods. Furthermore,~\ding{175}~\textbf{simply by substituting the original feature matrix with our augmented feature matrix, the performance of state-of-the-art generative pre-training methods can be further enhanced.} In Table~\ref{table:generative-results}, we observe that our method outperforms the SE variant in all cases. Even when compared with a strong baseline method (i.e., GAINT), GAugLLM prevails in 8 out of 10 scenarios, draws in 1, and falls behind in 1 scenarios. These results indicate that our mixture-of-prompt-expert technique can serve as an effective feature learning method in TAGs for graph generative models.

\begin{figure}[t]
\centering
\includegraphics[width=8cm]{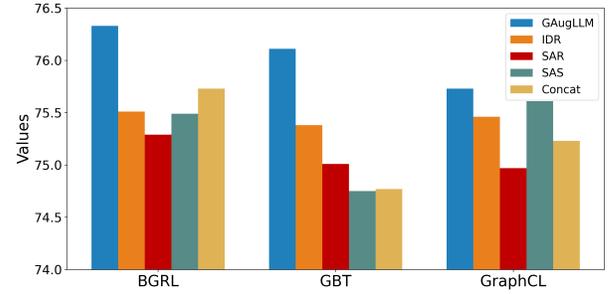}
\caption{Ablation study of GAugLLM on the History dataset. ``IDR'', ``SAR'', and ``SAS'' denote scenarios where we only employ the corresponding prompt expert for feature augmentation. ``Concat'' means we directly aggregate the hidden representations of all prompt experts as the final output.}
\label{fig:History-prompt-type}
\end{figure}

\begin{table}[t]
\centering
\caption{Ablation study of GAugLLM w.r.t. attention designs.}
\label{weight-strategies}
\begin{tabular}{l|l|l|l|l} 
\toprule
                        & Method                & BGRL       & GraphCL    & GBT         \\ 
\midrule
\multirow{2}{*}{PubMed} & w/o context          & 80.59±2.21 & 77.17±2.17 & 79.93±1.35  \\
                        & w/ context & 83.50±0.84 & 81.68±1.74 & 83.68±1.90  \\
\bottomrule
\end{tabular}
\vspace{-0.25cm}
\end{table}

\subsection{Ablation Study} 
\begin{figure}[t]
\centering
\includegraphics[width=8cm]{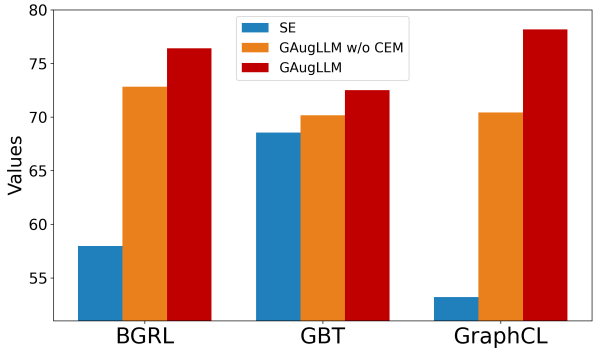}
\vspace{-0.25cm}
\caption{Ablation study of GAugLLM w.r.t. collaborative edge modifier on Photo dataset.}
\label{fig:Photo-Structure}
\end{figure}

To answer \textbf{RQ3}, we conduct a series of ablation studies to verify the contributions of different components in our model design. Specifically, we first test the impact of each individual prompt expert and reports the results in Figure~\ref{fig:History-prompt-type}. Then, we evaluate the contribution of the context-aware attention design in Eq.~\eqref{alo-atten3} in Table~\ref{weight-strategies}. Finally, we analyze the influence of the collaborative edge modifier in Figure~\ref{fig:Photo-Structure}. We make the following observations. 

~\ding{176}~\textbf{GAugLLM benefits from integrating multiple diverse prompt experts for feature augmentation.} As illustrated in Figure~\ref{fig:History-prompt-type}, GAugLLM consistently outperforms all four variants by a significant margin across three GCL backbones. Notably, even though both GAugLLM and the "Concat" variant utilize all prompt experts as input, GAugLLM outperforms "Concat" in all cases. The possible reason is that different nodes may prefer partial prompt experts for integrating the final augmented features. This comparison verifies our motivation to dynamically combine diverse prompt experts in a learnable way. 

\begin{table}[t]
\caption{The impact of different LLMs on GAugLLM.}
\label{LLM-backbones}
\centering
\begin{tabular}{l!{\vrule width \lightrulewidth}l!{\vrule width \lightrulewidth}l!{\vrule width \lightrulewidth}l!{\vrule width \lightrulewidth}l} 
\toprule
                         & Backbones    & BGRL       & GraphCL    & GBT         \\ 
\midrule
\multirow{3}{*}{PubMed}  & Mistral 8*7b & 83.50±0.84 & 81.68±1.74 & 83.68±1.90  \\
                         & ChatGPT-3.5       & 82.62±0.87 & 80.34±0.65 & 80.46±0.91  \\
                         & LLaMA2-13b   & 81.89±0.75 & 79.79±2.02 & 81.93±0.96  \\ 
\hline\hline
\multirow{3}{*}{History} & Mistral 8*7b & 76.33±0.88 & 75.11±0.4  & 76.11±0.4   \\
                         & ChatGPT-3.5       & 75.92±1.02 & 74.84±0.53 & 76.67±0.55  \\
                         & LLaMA2-13b   & 75.56±0.93 & 75.26±0.46 & 75.78±0.39  \\
\bottomrule
\end{tabular}
\end{table}

~\ding{177}~\textbf{By incorporating context information, GAugLLM provides an improved approach to integrating multiple prompt experts.} From Table~\ref{weight-strategies}, we can see that GAugLLM consistently generates more effective augmented features for state-of-the-art GCL methods. Notably, when the context-aware attention mechanism in Eq.~\eqref{alo-atten3} is not utilized, the performance of GAugLLM significantly declines. This outcome underscores the effectiveness of our proposed context-aware attention strategy in leveraging graph statistics.

~\ding{178}~\textbf{The proposed collaborative edge modifier scheme could significantly enhance the performance of GAugLLM compared to traditional masking strategies.} As depicted in Figure~\ref{fig:Photo-Structure}, we observe a substantial performance drop across three GCL methods when using the standard random edge masking for structure perturbation, whereas GAugLLM benefits significantly from the collaborative edge modifier. This comparison underscores the effectiveness of our proposed approach.

In addition to the main components, we also present an ablation study on the impact of different LLM backbones in Table~\ref{LLM-backbones}. From the table, we observe that~\ding{179}~\textbf{the performance gap between open-sourced and closed LLMs on GAugLLM is marginal.} In table~\ref{LLM-backbones}, we can see that GAugLLM performs generally much better on Mistral 8*7b and ChatGPT-3.5 compared with LLaMA2. More specifically, GAugLLM exhibits competitive or even superior performance on Mistral compared to ChatGPT. Since ChatGPT is a closed-sourced tool, this comparison validates the potential impact of our model in real-world scenarios as one can use the open-sourced LLM (i.e., Mistral 8*7b) without sacrificing performance.

\subsection{Sensitive Analysis}

\begin{figure}[t]
\centering
\includegraphics[width=8cm]{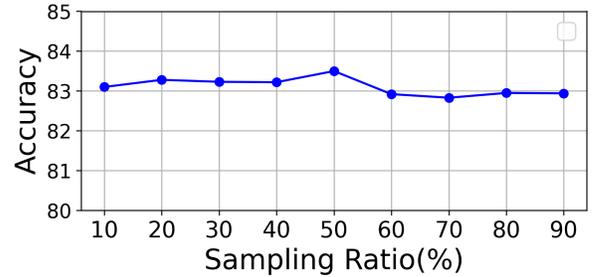}
\vspace{-0.2cm}
\caption{Sensitive analysis of GAugLLM w.r.t. the sampling ratio in collaborative edge modifier. }
\label{fig:pubmed-BGRL-sampling-ratio}
\end{figure}

To answer \textbf{RQ4}, we investigate the impact of different random sampling processes on GAugLLM. Specifically, we varied the sampling probability of the sample function in the collaborative edge modifier from 10\% to 90\% with a step size of 10\%. Figure~\ref{fig:pubmed-BGRL-sampling-ratio} reports the results. We observe that~\ding{180}~\textbf{The proposed collaborative edge modifier is robust to changes in the sampling ratio.} From Figure~\ref{fig:pubmed-BGRL-sampling-ratio}, we can see that GAugLLM performs the best when the sampling ratio is 50\%. We note that GAugLLM delivers very consistent accuracies across a wide range of sampling ratios, showing stability as the ratio increases from 10\% to 90\%, which would be desirable in real-world applications.

%% file: section/appendix.tex
\subsection{Experimental Configurations}
\label{appen-config}
For baselines, we report the baseline model results
based on their provided codes with official settings or results reported in previous researchse. If their settings or results are not available, we conduct a hyper-parameter search.
Table~\ref{config-GAugLLM-results} is the hyper-parameters for our own method GAugLLM in GCLs. Table~\ref{config-mix-experts} is the default setting for mix-of-prompt-expert module. One exception is that the epoch for arxiv is set to 1.

\begin{table}[H]
\centering
\caption{Configurations for each dataset on GCLs}
\label{config-GAugLLM-results}
\begin{tabular}{l!{\vrule width \lightrulewidth}l!{\vrule width \lightrulewidth}l!{\vrule width \lightrulewidth}l!{\vrule width \lightrulewidth}l} 
\toprule
                           & Setting        & BGRL  & GraphCL & GBT   \\ 
\midrule
\multirow{3}{*}{PubMed}    & lr             & 5e-4  & 1e-3    & 2e-3  \\
                           & encoder\_layer & 512   & 512     & 512   \\
                           & epoch          & 8000  & 1000    & 1000  \\ 
\hline\hline
\multirow{3}{*}{Arxiv}     & lr             & 1e-2  & 1e-3    & 1e-3  \\
                           & encoder\_layer & 512   & 256     & 256   \\
                           & epoch          & 1000  & 1000    & 1000  \\ 
\hline\hline
\multirow{3}{*}{History}   & lr             & 1e-3  & 5e-4    & 1e-4  \\
                           & encoder\_layer & 512   & 256     & 256   \\
                           & epoch          & 10000 & 5000    & 5000  \\ 
\hline\hline
\multirow{3}{*}{Photo}     & lr             & 1e-3  & 7e-4    & 5e-4  \\
                           & encoder\_layer & 512   & 256     & 256   \\
                           & epoch          & 10000 & 5000    & 5000  \\ 
\hline\hline
\multirow{3}{*}{Computers} & lr             & 1e-3  & 1e-3    & 1e-3  \\
                           & encoder\_layer & 512   & 256     & 256   \\
                           & epoch          & 10000 & 5000    & 5000  \\
\bottomrule
\end{tabular}
\end{table}

\begin{table}[h]
\centering
\caption{Default setting for mix-of-prompt-expert}
\label{config-mix-experts}
\begin{tabular}{c!{\vrule width \lightrulewidth}c} 
\toprule
Default Setting       &       \\ 
\hline
hidden\_dropout\_prob & 0.05  \\ 
\hline
batch\_size           & 32    \\ 
\hline
learning\_rate        & 6e-5  \\ 
\hline
epoch                 & 5/2/1     \\ 
\hline
attention~temperature & 0.2   \\
\bottomrule
\end{tabular}
\end{table}

\subsection{Prompt Expert Design}
\label{sec-feat-augmentor}

\begin{figure}[h!]
\centering
\includegraphics[width=0.4\textwidth]{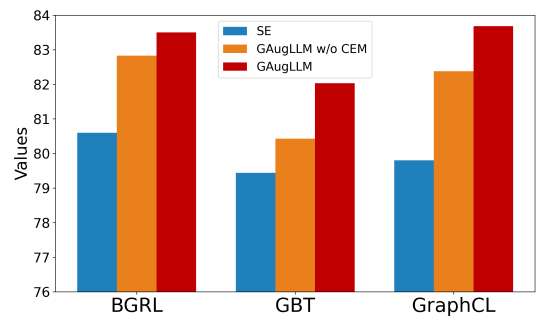}
\caption{Ablation study of GAugLLM w.r.t. collaborative edge modifier on PubMed dataset.}
\label{fig:PubMed-Structure}
\end{figure}

\begin{table*}[h]
\centering
\caption{Contex Prompt templates for different Experts.}
\label{feat-prompt-CPT}
\begin{tabular}{l!{\vrule width \lightrulewidth}l} 
\toprule
\textbf{Expert} & \textbf{Prompt Template}\\ 
\toprule
RAW             & This is the original text of this node.~The degree of this node is ... ... (Node information) \\ 
\hline\hline
IDR             & \begin{tabular}[c]{@{}l@{}}This is the explanation for classification based on the original text of this node.\\The degree of this node is ... We consider nodes with degree more than ... as head nodes. \\Head nodes have rich structure information in their connections with neighbor nodes.~\end{tabular} \\ 
\hline\hline
SAR             & \begin{tabular}[c]{@{}l@{}}This is the explanation for classification based on the original text with the understanding of its neighboring nodes.\\The degree of this node is ... We consider nodes with degree less than ... as tail nodes. \\Tail nodes have sparse structure information in their connections with neighbor nodes.~\textcolor[rgb]{0.639,0.082,0.082}{}\end{tabular}  \\ 
\hline\hline
SAS             & \begin{tabular}[c]{@{}l@{}}This is the summarization of the original text with the understanding of its neighboring nodes.\\The degree of this node is ... We consider degree less than ... and more than as mid nodes.~\end{tabular}\\
\bottomrule
\end{tabular}
\end{table*}

\begin{figure*}[h!]
\begin{center}
\includegraphics[scale=0.30]{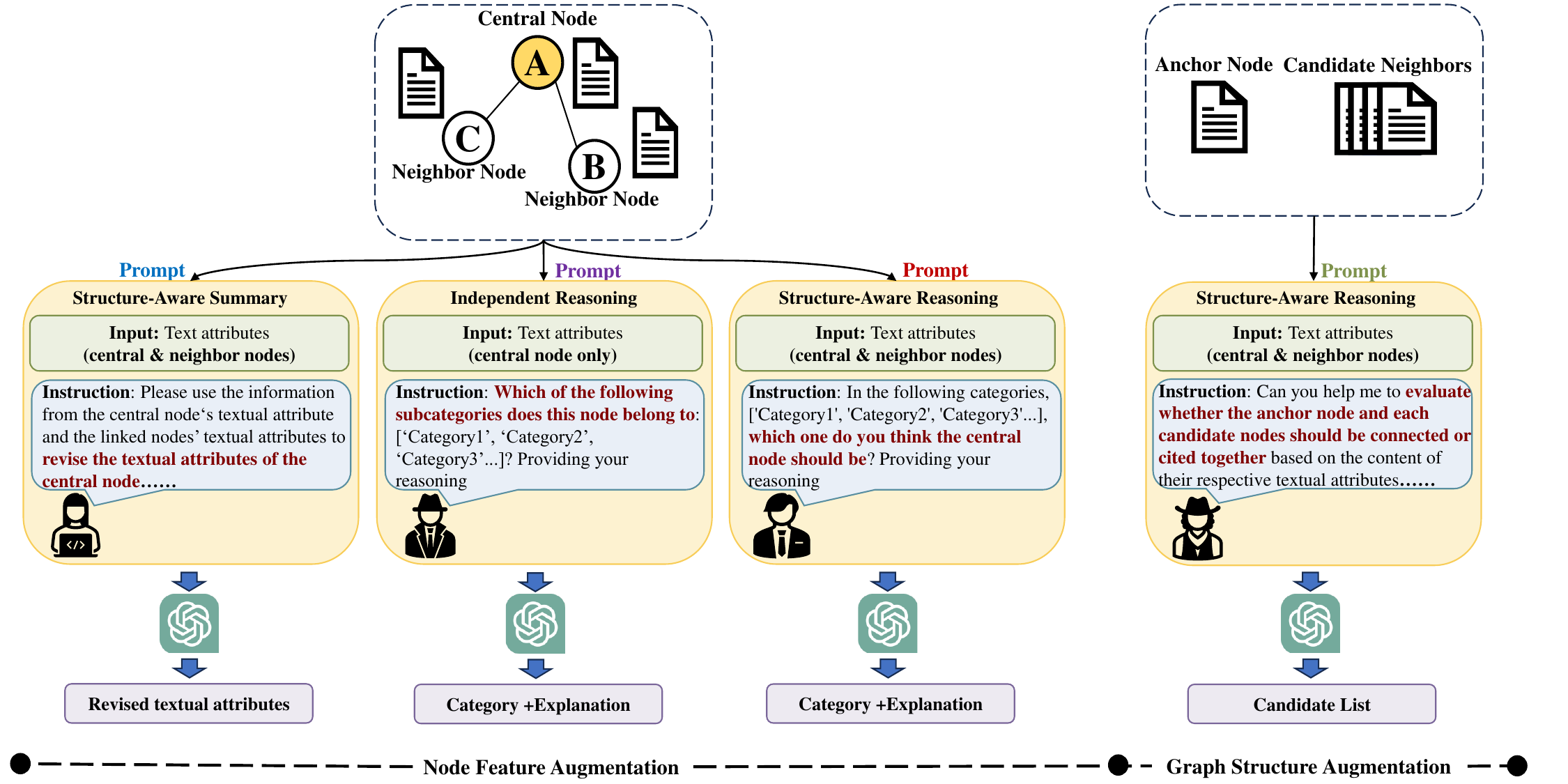}
\end{center}
\caption{LLM-as-GraphAugmentor. \textbf{Left}: LLMs are emloyed to perturb node features by influencing the input textual attributes. \textbf{Right}: LLMs are utilized to create new graph structures by modifying and adding edges between nodes. }
\label{graph-augmentation}
\end{figure*}

Given a node $v$ and its textual attribute $S_v$, traditional GCL methods typically create an augmented feature vector $\hat{\mathbf{x}}_v$ using purely stochastic functions, i.e., $\hat{\mathbf{x}}_v=\tau_f(\mathbf{x}_v)=\tau_f(\text{Emb}(S_v))$. However, this approach only introduces perturbations within the numerical space transformed by the $\text{Emb}(\cdot)$ module, which cannot effectively manipulate the original input textual attribute. To overcome this limitation, we propose to use LLMs to directly perturb the input text $S_v$ and obtain an augmented textual attribute $\hat{S}_v$ through three prompt templates (refer to Figure~\ref{graph-augmentation} (left)) outlined below.
\newpage

\noindent\textbf{Structure-Aware Summarization (SAS).} Let $\mathcal{S}_v^N=\{S_u|v\in \mathcal{N}_v\}$ represent the textual attribute set of node $v$'s neighbors. The idea of SAS is to query the LLM to create a summary of the anchor node $v$ by comprehending the semantic information from both its neighbors and itself. Specifically, for each node $v$, we construct a prompt that incorporates the textual attributes of the anchor node and its neighbors, denoted as $\{S_v, \mathcal{S}_v^N\}$, along with an instruction for revising its textual attribute. The general prompt format is illustrated in the left panel of Figure~\ref{graph-augmentation} (left). 
Finally, we employ these summarized textual attributes to represent the augmented attribute $\hat{S}_v$.

\noindent\textbf{Independent Reasoning (IDR).} In contrast to SAS, which concentrates on text summarization, IDR adopts an ``open-ended" approach when querying the LLM. This entails instructing the model to make predictions across potential categories and to provide explanations for its decisions. The underlying philosophy here is that such a reasoning task will prompt the LLM to comprehend the semantic significance of the input textual attribute at a higher level, with an emphasis on the most vital and relevant factors~\citep{he2023explanations}. Following this principle, for each node $v$, we generate a prompt that takes the textual attribute of the anchor node as input and instructs the LLM to predict the category of this node and provide explanations. The general prompt format is illustrated in the middle panel of Figure~\ref{graph-augmentation} (left). 
We utilize the prediction and explanations to represent the augmented attribute $\hat{S}_v$.   

\noindent\textbf{Structure-Aware Reasoning (SAR).} Taking a step beyond IDR, SAR integrates structural information into the reasoning process. The rationale for this lies in the notion that connected nodes can aid in deducing the topic of the anchor node. Specifically, for each node $v$, we devise a prompt that encompasses the textual attributes of the anchor node $S_v$ and its neighbors $S_v^N$, along with an open-ended query concerning the potential category of the node. The general prompt format is given in the right panel of Figure~\ref{graph-augmentation} (left). 
Similar to IDR, we employ the prediction and explanations to denote the augmented attribute $\hat{S}_v$.

To reduce the query overhead of ChatGPT, we randomly sample 10 neighbors for each anchor node in structure-aware prompts (i.e., SAS and SAR) in our experiments.

\subsection{Collaborative Edge Modifier}
\label{appen-edge}
This section is dedicated to elucidating the algorithm behind the Collaborative Edge Modifier. The algorithm operates in two distinct phases. Initially, in the first phase, we deploy a Language Model (LLM) to generate two sets of edges. Subsequently, in the second phase, we proceed to either incorporate or eliminate portions of the graph structure based on the edges produced in the initial phase. For those interested in the finer details, the pseudocode for this process is provided in Algorithm~\ref{algo-cem}.
\begin{algorithm}[h]
\caption{Collaborative Edge Modifier}
\begin{algorithmic}[1]
\Procedure{Structure\_Augmentation}{$G, v, A, LLM$}
    \label{algo-cem}
    \State \textit{// First stage: structure-aware top candidate generation.}
    
    \State $N_v \gets$ \text{ConnectedNodes}(v)
    \State $\bar{N_v} \gets$ \text{DisconnectedNodes}(v)
    \State
    \State \textit{//network embedding algorithm}
    \State $\mathcal{E}_v^{\text{spu}} \gets$ \text{SelectTopK}($N_v$, 10, descending)
    \State $\mathcal{E}_v^{\text{mis}} \gets$ \text{SelectTopK}($\bar{N_v}$, 10, ascending)
    
    \State prompt\_connect $\gets$ \text{CreatePrompt}$(v, \mathcal{E}_v^{\text{spu}})$
    \State prompt\_disconnect $\gets$ \text{CreatePrompt}$(v, \mathcal{E}_v^{\text{mis}})$
    \State candidates\_discard $\gets$ \text{LLM}(prompt\_connect)
    \State candidates\_add $\gets$ \text{LLM}(prompt\_disconnect)
    \State 
    \State \textit{//Second Stage: Update adjacency matrix based on LLM decisions with a certain accept rate.}
    \For{each epoch in contrastive training}
        \For{each node $u$ in v}
           \State edges\_add $\gets$ \text{RandomSelect}$(u, candidates\_add, 0.5)$
           \State edges\_discard $\gets$ \text{RandomSelect}\\
            \hskip\algorithmicindent $(u, candidates\_discard, 0.5)$
           \State Update $\hat{A}[v][u]$ with edges\_add and edges\_discard
        \EndFor
        \State Use {A} and $\hat{A}$ for contrastive training
    \EndFor
\EndProcedure
\end{algorithmic}
\end{algorithm}